%% file: main.tex
\documentclass[10pt,conference]{IEEEtran}

\IEEEoverridecommandlockouts
\usepackage{cite}
\usepackage{amsmath,amssymb,amsfonts}
\usepackage{algorithmic}
\usepackage{graphicx}
\usepackage{textcomp}
\usepackage{subcaption}
\usepackage{pgfplots}
\usepackage[printonlyused,nolist]{acronym}
\usepackage{adjustbox} 
\usepackage{multirow} 
\usepackage{url}
\usepackage{listings}
\input{lstlistings_javascript}

\usepackage{booktabs}
\usepackage{wrapfig}
\newtheorem{definition}{Definition}
\newcommand{\code}[1]{\texttt{#1}}
\newcommand{\scode}[1]{\texttt{\small #1}}

\newcommand{\name}[0]{IdBench}

\begin{document}

\title{\name{}: Evaluating Semantic Representations of Identifier Names in Source Code
\thanks{This work was supported by the European Research Council (ERC, grant agreement 851895), and by the German Research Foundation within the ConcSys and Perf4JS projects.}}

\author{\IEEEauthorblockN{Yaza Wainakh}
\IEEEauthorblockA{\textit{Department of Computer Science} \\
\textit{TU Darmstadt}\\
Darmstadt, Germany \\
yaza.wainakh@gmail.com}
\and
\IEEEauthorblockN{Moiz Rauf}
\IEEEauthorblockA{\textit{Department of Computer Science} \\
\textit{University of Stuttgart}\\
Stuttgart, Germany \\
moiz.rauf@iste.uni-stuttgart.de}
\and
\IEEEauthorblockN{Michael Pradel}
\IEEEauthorblockA{\textit{Department of Computer Science} \\
\textit{University of Stuttgart}\\
Stuttgart, Germany \\
michael@binaervarianz.de}
}

\maketitle

\begin{abstract}
Identifier names convey useful information about the intended semantics of code.
Name-based program analyses use this information, e.g., to detect bugs, to predict types, and to improve the readability of code.
At the core of name-based analyses are semantic representations of identifiers, e.g., in the form of learned embeddings.
The high-level goal of such a representation is to encode whether two identifiers, e.g., \code{len} and \code{size}, are semantically similar.
Unfortunately, it is currently unclear to what extent semantic representations match the semantic relatedness and similarity perceived by developers.
This paper presents \name{}, the first benchmark for evaluating semantic representations against a ground truth created from thousands of ratings by 500 software developers.
We use \name{} to study state-of-the-art embedding techniques proposed for natural language, an embedding technique specifically designed for source code, and lexical string distance functions.
Our results show that the effectiveness of semantic representations varies significantly and that the best available embeddings successfully represent semantic relatedness.
On the downside, no existing technique provides a satisfactory representation of semantic similarities, among other reasons because identifiers with opposing meanings are incorrectly considered to be similar, which may lead to fatal mistakes, e.g., in a refactoring tool.
Studying the strengths and weaknesses of the different techniques shows that they complement each other.
As a first step toward exploiting this complementarity, we present an ensemble model that combines existing techniques and that clearly outperforms the best available semantic representation.
\end{abstract}

\section{Introduction}

Identifier names play an important role in writing, understanding, and maintaining high-quality source code~\cite{Butler2010}.
Because they convey information about the meaning of variables, functions, classes, and other program elements, developers often rely on identifiers to understand code written by themselves and others.
%
Beyond developers, various automated techniques analyze, use, and improve identifier names.
For example, identifiers have been used to find programming errors~\cite{issta2011,oopsla2017,oopsla2018-DeepBugs,DBLP:conf/sigsoft/KateOZEX18}, to mine specifications~\cite{Zhong2009a}, to infer types~\cite{Xu2016,icse2019}, to predict the name of a method~\cite{Allamanis2015}, or to complete partial code using a learned language model~\cite{DBLP:conf/acl/ChangPCR18}.
Techniques for improving identifier names pinpoint inappropriate names~\cite{Host2009} and suggest more suitable names~\cite{Liu2019}.
The basic idea of all these approaches is to infer the intended meaning of a piece of code from the natural language information in identifiers, possibly along with other information, such as the structure of code, data flow, and control flow.
We here refer to program analyses that rely on identifier names as a primary source of information as \emph{name-based analyses}.

Most name-based analyses reason about names in one of two ways.
First, some approaches build upon \emph{string distance functions}, such as the Levenshtein distance, sometimes in combination with algorithms for tokenizing names, e.g., based on underscore or camel-case notation~\cite{Butler2011}.
Given a pair of identifiers, e.g. \code{len} and \code{length}, a string distance function yields a real-valued number that indicates to what extent the character sequences in the identifiers resemble each other.
String distance functions are at the core of name-based analyses to detect name-related bugs~\cite{issta2011,oopsla2017}, to predict types~\cite{Xu2016}, to improve identifier names~\cite{Jiang2018}, or to suggest appropriate names~\cite{icse2016-names}.
Second, another approach, which has become popular more recently, are neural network-learned \emph{embeddings of identifiers}.
An embedding maps each identifier into a continuous vector representation, so that similar identifiers are mapped to similar vectors.
Embeddings implicitly define a similarity function via the cosine similarity of embedding vectors.
For example, embeddings of identifiers are at the core of neural program analyses~\cite{NeuralSoftwareAnalysis} to predict types~\cite{icse2019}, to detect bugs~\cite{oopsla2018-DeepBugs}, to de-obfuscate code~\cite{alon2018general}, to complete partial code~\cite{DBLP:conf/acl/ChangPCR18}, and to map API elements across programming languages~\cite{Nguyen2017}.

The common aim of both string distance functions and embeddings of identifiers is to reason about the semantics of identifiers, and we hence call both of them \emph{semantic representations of identifiers}, or short \emph{semantic representations}.
The overall effectiveness of a name-based analysis relies on the assumption that the underlying semantic representation encodes some kind of semantic relationship between identifiers.
For example, two semantically similar identifiers, such as \code{len} and \code{length}, should be closer to each other than two unrelated identifiers, such as \code{length} and \code{click}.

%
%

Despite the importance of semantic representations for name-based analyses, it is currently unclear how well existing approaches actually represent semantic relationships.
Specifically, we are interested in the following questions:

\paragraph{RQ~1: How accurately do state-of-the-art semantic representations match the \emph{semantic relatedness} of identifiers as perceived by software developers?}
``Relatedness'' here means the degree of association between two identifiers, which covers various possible relations between them, e.g., being used in the same application domain or being opposites of each other.
For example, \code{top} and \code{bottom} are related because they are opposites, \code{click} and \code{dblclick} are related because they belong to the same general concept, and \code{getBorderWidth} and \code{getPadding} are related because they belong to the same application domain. 
The relatedness of identifiers is relevant for tools that reason about the broad meaning of code elements, e.g., to predict the types of functions~\cite{icse2019,Hellendoorn2018}.

\paragraph{RQ~2: How accurately do state-of-the-art semantic representations match the \emph{semantic similarity} of identifiers as perceived by software developers?}
``Similarity'' here means the degree to which two identifiers have the same meaning, in the sense that one could substitute the other without changing the overall meaning~\cite{miller1991contextual}.
For example, \code{length} and \code{size}, as well as \code{username} and \code{userid}, are similar to each other.
The similarity of identifiers is, e.g., relevant for name-based bug detection tools~\cite{oopsla2018-DeepBugs,Allamanis2017b}.

\paragraph{RQ~3: What are the strengths and weaknesses of the existing semantic representations?} 
Better understanding why particular techniques sometimes succeed or fail to accurately represent identifiers will enable improving the current semantic representations.

\paragraph{RQ~4: Do the existing semantic representations complement each other?}
If current techniques are complementary, it may be possible to combine them in a way that outperforms the individual techniques.

\smallskip
\noindent

Addressing these questions relies on a way to measure and compare the effectiveness of semantic representations of identifiers in source code.
This paper presents \name{}, the first benchmark for this task, which is based on a dataset of developer assessments about the relatedness and similarity of pairs of identifiers.
We gather this dataset through surveys that show real-world identifiers and code snippets to hundreds of developers, asking them to rate their semantic relationship.
Taking the developer assessments as a gold standard, \name{} allows for evaluating semantic representations in a systematic way by measuring to what extent a semantic representation agrees with ratings given by developers.
Moreover, inspecting pairs of identifiers for which a representation strongly agrees or disagrees with the benchmark helps understand the strengths and weaknesses of the representation.


Applying our methodology to seven widely used semantic representations leads to various novel insights.
We find that different techniques differ heavily in their ability to accurately represent identifier relatedness and similarity.
The best among the studied techniques, the CBOW variant of FastText~\cite{Bojanowski2017}, accurately represents the relatedness of identifiers (RQ~1), but none of the available techniques accurately represents the similarity of identifiers (RQ~2).
Studying the strengths and weaknesses of each technique (RQ~3) shows that some embeddings are confused about identifiers with opposite meaning, e.g., \code{rows} and \code{cols}, about identifiers that belong to the same application domain but are not similar, and about synonyms, e.g., \code{file} and \code{record}.
Furthermore, practically all techniques struggle with identifiers that use abbreviations, which are very common in software.
We also find that simple string distance functions, which measure the similarity of identifiers without any learning, are surprisingly effective, and even outperform some learned embeddings for the similarity task.

A close inspection of the results shows that different techniques complement each other (RQ~4).
To benefit from the strengths of multiple techniques, we present a new semantic representation that combines the available techniques into an ensemble model based on features of identifiers, such as the number of characters or whether an identifier contains non-dictionary words.
The ensemble model clearly outperforms each of the existing semantic representations, improving agreement with developers by 6\% and 19\% for relatedness and similarity, respectively.


In summary, this paper makes the following contributions.
\begin{itemize}
\item \emph{A reusable benchmark}.
We make available a benchmark of hundreds of pairs of identifiers, providing a way to systematically evaluate existing and future embeddings.\footnote{\url{https://github.com/sola-st/IdBench}}
To the best of our knowledge, this is the first benchmark to systematically evaluate semantic representations of identifiers.
\item \emph{Novel insights}.
Our study reveals both strengths and limitations of current semantic representations, along with concrete examples to illustrate them.
These insights provide a basis for future work on better semantic representations.
\item \emph{A technique that outperforms the state-of-the-art}.
Combining the currently available techniques based on a few simple features yields a semantic representation that clearly outperforms all individual techniques.
\end{itemize}

\section{Methodology}
\label{sec:benchmark}

To measure and compare the accuracy of semantic representations, we gather thousands of ratings from 500 developers (Section~\ref{sec:surveys}).
Cleaning and compiling this raw dataset into a benchmark yields several hundreds of pairs of identifiers with gold standard similarities (Section~\ref{sec:cleaning}).
We then measure the agreement between the gold standard and state-of-the-art semantic representations (Section~\ref{sec:measuring agreement}), where we study two string distance functions and five learned embeddings (Section~\ref{sec:semantic representations}).
We apply our methodology to JavaScript code, because recent work on identifier names and code embeddings focuses on this language~\cite{oopsla2018-DeepBugs,alon2019code2vec,alon2018general,icse2019}, but our methodology can also be applied to other languages.

\subsection{Developer Surveys}
\label{sec:surveys}

\name{} includes three benchmark tasks: A \emph{relatedness task} and two tasks to measure how well an embedding reflects the similarity of identifiers: a \emph{similarity task} and a \emph{contextual similarity task}. The following describes how we gather developer assessments that provide data for these tasks.
The supplementary material provides additional examples and details of the survey setup.

\paragraph{Direct Survey of Developer Assessments}
\label{sec:direct survey}

\begin{figure}
	\begin{subfigure}{\linewidth}
		Identifiers: \code{radians}, \code{angle}
		
		\vspace{.7em}
		1) \emph{How related are the identifiers?}
		
		\vspace{.4em}
		\begin{tabular}{@{}p{7em}cccccp{6em}@{}}
			Unrelated & $\bigcirc$ & $\bigcirc$ & $\bigcirc$ & $\bigcirc$ & $\bigcirc$ & Related \\
		\end{tabular}

		\vspace{1em}
		2) \emph{Could one substitute the other?}

		\vspace{.4em}
		\begin{tabular}{@{}p{7em}cccccp{6em}@{}}
			Not substitutable & $\bigcirc$ & $\bigcirc$ & $\bigcirc$ & $\bigcirc$ & $\bigcirc$ & Substi\-tutable \\
		\end{tabular}
		\caption{Direct survey.}
		\label{fig:direct survey}
	\end{subfigure}
	\vspace{2em}
	
	\begin{subfigure}{\linewidth}
		\emph{Which identifier fits best into the blanks?}
		
		\vspace{.8em}
		$\bigcirc$ \small{\code{positions}}
		\hspace{1.5em}
		$\bigcirc$ \small{\code{indices}}
		
		\vspace{.5em}
\begin{lstlisting}[numbers=none,basicstyle=\ttfamily\small]
Opentip./*#\HL#*/____/*#\HLoff#*/ = [ "top", "topRight",
    "right", "bottomRight", "bottom",
    "bottomLeft", "left", "topLeft"];
Opentip.position = {};
_ref = Opentip./*#\HL#*/____/*#\HLoff#*/;
\end{lstlisting}
		\caption{Indirect survey.}
		\label{fig:indirect survey}
	\end{subfigure}
	\caption{Examples of the developer surveys.}
	\label{fig:surveys}
\end{figure}

This survey shows two identifiers to a developer and then directly asks how related and how similar the identifiers are.
Figure~\ref{fig:direct survey} shows an example question from the survey.
The developer is shown pairs of identifiers and is then asked to rate on a five-point Likert scale how related and how similar these identifiers are to each other.
In total, each developer is shown 18 pairs of identifiers, which we randomly sample from a larger pool of pairs.
Before showing the questions, we provide a brief description of what the developers are supposed to do, including an explanation of the terms ``related'' and ``substitutable''.
The ratings gathered in the direct survey are the basis for the relatedness task and the similarity task of \name{}.

\paragraph{Indirect Survey of Developer Assessments}
\label{sec:indirect survey}

This survey asks developers to pick an identifier that best fits a given code context, which indirectly asks about the similarity of identifiers.
The motivation is that identifier names alone may not provide enough information to fully judge how similar they are~\cite{miller1991contextual}.
For example, without any context, identifiers \code{idx} and \code{hl} may cause confusion for developers who are trying to judge their similarity. 
The survey addresses this challenge by showing the code context in which an identifier occurs, and by asking the developers to decide which of two given identifiers best fits this context.
If, for a specific pair of identifiers, the developers choose both identifiers equally often, then the identifiers are likely to be similar to each other, since one can substitute the other.
Figure~\ref{fig:indirect survey} shows a question asked during the indirect survey.
As shown in the example, for code contexts where the identifier occurs multiple times, we show multiple blanks that all refer to the same identifier.
In total, we show 15 such questions to each participant of the survey, where the 15 identifier pairs are randomly selected from the set of all studied pairs.
The ratings gathered in the indirect survey are the basis for the contextual similarity tasks of \name{}.


\paragraph{Selection of Identifiers and Code Examples}
\label{sec:selection}

We select identifiers and code contexts from a corpus of 50,000 JavaScript files~\cite{raychev2016learning_150KJavaScript}.
We select 300 pairs, made out of 488 identifiers, through a combination of automated and manual selection, aimed at a diverse set that covers different degrees of similarity and relatedness.
At first, we extract from the code corpus all identifier names that appear more than 50 times, including method names, variable names, property names, and other types of identifiers.
A naive approach would be to randomly sample pairs among those identifiers.
However, this naive approach would result almost only in unrelated and dissimilar identifier pairs.
Instead, we follow a methodology proposed for natural language~\cite{MEN}, which ranks all pairs based on the cosine similarity according to a given embedding, and then selects pairs from different ranges in the ranking.
We select pairs using two embeddings~\cite{word2veca,alon2018general}. 
The fact that these embeddings are later also evaluated with the benchmark does not introduce bias because the ground truth of the benchmark is constructed only from the human ratings, not from the embeddings.
In addition to pairs selected as suggested in~\cite{MEN}, we manually select some synonym pairs, which we observed to lack otherwise, and add randomly selected pairs, which are likely to be unrelated.
The manual selection was done before evaluating any semantic representations to avoid biasing the benchmark.

To gather the code contexts for the indirect survey, we search the code corpus for occurrences of the selected identifiers.
As the size of the context, we choose five lines, aiming to provide sufficient context to pick the best identifier without overwhelming the study participants with large amounts of code.
For each identifier, we randomly select five different contexts. 
When showing a specific pair of identifiers to a developer, we randomly select one of the gathered contexts for one of the two identifiers.

\begin{table}
	\caption{Occurrences of \name{} identifiers in code corpora of different languages.}
	\label{tab:PL50k}
	\setlength{\tabcolsep}{6.5pt}
	\begin{tabular}{@{}lrr|rrr@{}}
		\toprule
		& \multicolumn{2}{c}{Total occurrences} & \multicolumn{3}{c}{Occurrences\ of individual\ identifiers} \\
		\cmidrule{2-3}
		\cmidrule{4-6}
		Language & Number & Perc. & Min & Mean & Max \\
		\midrule
		JavaScript & 3,697,498 & 12.5\% & 62 & 7,639 & 629,413  \\
		Python & 2,279,866 & 14.8\% & 0 & 4,710 & 1,367,832 \\
		Java & 757,064 & 6.3\% & 0 & 1,564 & 119424 \\
		\bottomrule
	\end{tabular}
\end{table}

Table~\ref{tab:PL50k} shows how often the selected identifiers occur in the JavaScript corpus.
Overall, the identifiers in \name{} occur 3.7 million times, which covers 12.5\% of all identifier occurrences. 
Even though this was not a criterion when selecting the identifiers, the benchmarks covers a non-negligible portion of real-world code.
The table also shows how often individual identifiers occur, which is 7,639 times, on average.
To assess whether \name{} could also be used for other languages, we also measure the occurrences in Python~\cite{Raychev2016a} and Java code corpora~\cite{Allamanis2013} with 50,000 files each.
As shown in Table~\ref{tab:PL50k}, the identifiers are also frequent in code beyond JavaScript, with an average number of occurrences of 4,710 and 1,564 in the Python and Java corpora, respectively.
A manual analysis shows that identifiers that occur across languages cover general programming terminology, whereas identifiers that appears in JavaScript only are mostly specific to the web domain, e.g., \code{tag\_h4} or \code{DomRange}.

To better understand whether \name{} covers identifiers that appear in different syntactic roles, we measure for each identifier how often it used as a function name, variable name, or property name.
We then assign each identifier to one of these roles based on whether the majority of its occurrences is in a specific role.
The measurements show that 17\% of the identifiers are primarily function names, 18\% are primarily variables names, 34\% are primarily property names, and the rest is commonly used in multiple roles.

\paragraph{Participants}
We recruit developers to participate in the survey in several ways.
About half of the participants are volunteers recruited via personal contacts, posts in public developer forums, and a post in an internal forum within a major software company.
The other half of the participants were recruited via Amazon Mechanical Turk, where we offered a compensation of one US dollar for completing both surveys.
On average, participants took around 15 minutes to complete both surveys.
That is, the offered compensation matches the average salary of software developers in some countries of the world.\footnote{\url{https://www.payscale.com/research/IN/Job=Software_Developer/Salary}}
In total, 500 developers participate in the survey.
Most participants live in North America and in India, and they have at least five years of experience in software development.

\subsection{Data Cleaning}
\label{sec:cleaning}

Crowd-sourced surveys may contain noise, e.g., due to lack of expertise or involvement by the participants~\cite{kittur2008crowdsourcing}.
To address this challenge, we gather at least ten ratings per pair of identifiers and then clean the data based on the inter-rater agreement, which has been found effective in other crowd-sourced surveys~\cite{nowak2010reliable}.

\paragraph{Removing Outlier Participants}
As a first filter, we remove outlier participants based on the inter-rater agreement, which measures the degree of agreement between participants.
We use Krippendorf's alpha coefficient, because it handles unequal sample sizes, which fits our data, as not all participants rate the same pairs and not all pairs have the same number of ratings.
The coefficient ranges between zero and one, where zero represents complete disagreement and one represents perfect agreement. 
For each participant, we calculate the difference between her rating and the average of all the other ratings for each pair.
Then, we average these differences for each rater, and discard participants with a difference above a threshold~$\tau$ (values given in Table~\ref{tab:benchmark size}).
We perform this computation both for the relatedness and similarity ratings from the direct survey, and then remove outliers based on the average difference across both ratings.

\paragraph{Removing Downer Participants}
As a second filter, we eliminate participants that decrease the overall inter-rater agreement (IRA).
We call such participants \emph{downers}~\cite{hill2015simlex}, because they bring the agreement level between all participants down.
For each participant $p$, we compute IRA\textsubscript{sim} and IRA\textsubscript{rel} before and after removing $p$ from the data. 
If IRA\textsubscript{sim} or IRA\textsubscript{rel} increases by at least 10\%, then we discard that participant's ratings.


\paragraph{Removing Outlier Pairs}
As a third filter, we eliminate some pairs of identifiers used in the indirect survey.
Since our random selection of code contexts may include contexts that are not helpful in deciding about the most suitable identifier, the ratings for some pairs may be misleading.
For example, this is the case for code contexts that contain short and meaningless identifiers or that mostly consist of comments unrelated to the missing identifier.
To mitigate this problem, we remove a pair if the difference in similarity as rated in the direct and indirect surveys exceeds some threshold~$\theta$ (values given in Table~\ref{tab:benchmark size}).

\begin{table}
	\caption{Benchmark sizes and inter-rater agreement (IRA).}
	\label{tab:benchmark size}
	\setlength{\tabcolsep}{5.5pt}
	\begin{tabular}{@{}llr|rr|rr|c@{}}
		\toprule
		Size &
		\multicolumn{2}{c}{Thresholds} &
		\multicolumn{5}{c}{Task} \\
		\cmidrule{2-8}
		& $\tau$ & $\theta$ &
		\multicolumn{2}{c}{Relatedness} &
		\multicolumn{2}{c}{Similarity} &
		Contextual\ simil. \\
		\cmidrule{4-8}
		&&& Pairs & IRA & Pairs & IRA & Pairs \\
		\midrule
		Small & 0.215 & 0.4 & 167 & 0.67 & 167 & 0.62 & 115 \\
		Medium & 0.23 & 0.5 & 247 & 0.61 & 247 & 0.57 & 145 \\
		Large & 0.25 & 0.6 & 291 & 0.56 & 291 & 0.51 & 176 \\
		\bottomrule		
	\end{tabular}
\end{table}
	
\begin{table}
	\caption{Pairs of identifiers with their gold standard similarities.}
	\label{tab:benchmark examples}
	\setlength{\tabcolsep}{4pt}
	\begin{tabular}{@{}llrrr@{}}
		\toprule
		&& \multicolumn{3}{c}{Score} \\
		\cmidrule{3-5}
		Identifier 1 & Identifier 2 & Related- & Similar- & Contextual \\
		&&ness&ity& similarity \\
		\midrule
		\scode{substr}&\scode{substring}&0.94&1.00&0.89\\
		\scode{setMinutes}&\scode{setSeconds}&0.91&0.22&0.06\\
		\scode{reset}&\scode{clear}&0.90&0.89&0.94\\
		\scode{rows}&\scode{columns}&0.88&0.08&0.22\\
		\scode{setInterval}&\scode{clearInterval}&0.86&0.09&0.34\\
		\scode{count}&\scode{total}&0.83&0.81&0.79\\
		\scode{item}&\scode{entry}&0.78&0.77&0.92\\
		\scode{miny}&\scode{ypos}&0.68&0.37&0.02\\
		\scode{events}&\scode{rchecked}&0.16&0.14&0.18\\
		\scode{re}&\scode{destruct}&0.06&0.02&0.02\\
		\bottomrule
	\end{tabular}
\end{table}


\vspace{1em}
\noindent
Table~\ref{tab:benchmark size} shows the number of identifier pairs that remain in the benchmark after data cleaning.
For each of the three tasks, we provide a \emph{small}, \emph{medium}, and \emph{large} benchmark, which differ in the thresholds used during data cleaning.
The smaller benchmarks use stricter thresholds and hence provide higher agreements between the participants, whereas the larger benchmarks offer more pairs.
The thresholds are selected to strike a balance between increasing the overall inter-rater agreement while keeping enough pairs and ratings to form a representative benchmark.

\subsection{Measuring Agreement with the Benchmark}
\label{sec:measuring agreement}

Given the ground truth similarities and a semantic representation technique, we want to measure to what extent both agree with each other.

\paragraph{Converting Ratings to Scores}
As a first step of measuring the agreement with the benchmark, we convert the ratings gathered for a specific pair during the developer surveys into a similarity score in the $[0,1]$ range.
For the direct survey, we scale the 5-point Likert-scale ratings into the $[0,1]$ range and average all ratings for a specific pair of identifiers.
For the indirect survey, we use a signal detection theory-based approach for converting the collected ratings into numeric values, which has been previously used to create a similarity benchmark for natural languages \cite{miller1991contextual}.
This conversion yields an unbounded distance measure $d$ for each pair, which we convert into a similarity score $s$ by normalizing and inverting the distance:
$s = 1 - \frac{d - min_d}{max_d - min_d}$
where $min_d$ and $max_d$ are the minimum and maximum distances across all pairs.

\paragraph{Examples}
Table~\ref{tab:benchmark examples} shows representative examples of identifier pairs and their scores for the three benchmark tasks.\footnote{The full list of identifiers pairs is available for download as part of our benchmark.}
The examples illustrate that the scores match human intuition and that the gold standard clearly distinguishes relatedness from similarity.
Some of the highly related and highly similar pairs, e.g., \code{substr} and \code{substring}, are lexically similar, while others are synonyms, e.g., \code{count} and \code{total}.
The identifiers \code{rows} and \code{columns} are strongly related, but one cannot substitute the other, and they hence have low similarity.
Similarly \code{miny}, \code{ypos} represent distinct properties of the variable \code{y}, which is why they are related but not similar. 
Finally, some pairs are either weakly or not at all related, e.g., \code{re} and \code{destruct}.

\paragraph{Correlation with benchmark}
We measure the magnitude of agreement of a semantic representation with \name{} by computing Spearman's rank correlation between the similarities of pairs of identifier vectors according to the semantic representation and our gold standard of similarity scores.

\begin{definition}[Correlation with benchmark]
Given $n$ pairs $(s_i,g_i)$ of similarity scores, where $s_i$ is computed by a semantic representation and $g_i$ is the gold standard, let $rank(s_i)$ and $rank(g_i)$ be the ranks of $s_i$ and $g_i$, respectively.
The correlation of the semantic representation with the benchmark is
$r = \dfrac{cov(rank(s_i), rank(g_i))}{\sigma_{rank(s_i)} \sigma_{rank(g_i)}}$
where $cov$ and $\sigma$ are covariance and standard deviation of the rank variables, respectively.
\end{definition}

The correlation ranges between 1 (perfect agreement) and -1 (complete disagreement).
For string distance functions, we compute the similarity score $s_i=1-d_{norm}$ for each pair based on a normalized version $d_{norm}$ of the distance returned by the string distance function.
We use Spearman's rank correlation because directly comparing absolute similarities across different embeddings may be misleading~\cite{zhelezniak2019correlation}.
The reason is that, depending on how ``wide'' or ``narrow'' an embedding space is, a cosine similarity of 0.3 may mean a rather high or a rather low similarity.
A rank-based comparison, as provided by Spearman's rank correlation, is more robust to different ways of populating the embedding space with identifiers than computing the correlation of absolute similarities.

\subsection{Embeddings and String Distance Functions}
\label{sec:semantic representations}

To assess how accurately existing semantic representations encode the relatedness and similarity of identifiers, we evaluate seven semantic representations against \name{}: Two string distance functions and five learned embeddings.

String distance functions use lexical similarity as a proxy for the semantic relatedness of identifiers.
We consider these functions because they are used in name-based bug detection tools~\cite{issta2011}, including a bug detection tool deployed at Google~\cite{oopsla2017},
to improve identifier names~\cite{Jiang2018},
and to suggest appropriate names~\cite{icse2016-names}.
The two string distance functions we evaluate are:
\begin{itemize}
  \item ``LV'': \emph{Levenshtein's} edit distance, which is the number of character insertions, deletions, and substitutions required to transform one identifier into another.
  \item ``NW'': \emph{Needleman-Wunsch} distance~\cite{needleman1970general}, which generalizes the Levenshtein distance by computing global alignments of two strings. 
\end{itemize}

Learned embeddings are popular in recent name-based analyses, e.g., for bug detection~\cite{oopsla2018-DeepBugs}, type prediction~\cite{icse2019}, and for predicting names and types of program elements~\cite{alon2018general}.
The five learned embeddings we evaluate are:
\begin{itemize}
\item ``w2v-cbow'': The continuous bag of words variant of \emph{Word2vec}~\cite{word2veca,word2vecb}.

\item ``w2v-sg'': The skip-gram variant of \emph{Word2vec}.

\item ``FT-cbow'': The continuous bag of words variant of \emph{FastText}~\cite{Bojanowski2017}, a sub-word extension of Word2vec that represents words as character n-grams.

\item ``FT-sg'': The skip-gram variant of \emph{FastText}.

\item ``path-based'': An embedding technique specifically designed for code, which learns from paths through a structural, tree-based representation of code~\cite{alon2018general}.
\end{itemize}

We train all embeddings on the same code corpus of 50,000 JavaScript files~\cite{raychev2016learning_150KJavaScript}. For each embedding, we experiment with various hyper-parameters (e.g., dimension, number of context words) and report results only for the best performing models.\footnote{Details on the hyperparameters and how we tuned them are available in the supplementary material.}
We provide all identifiers as they are to the semantic representations, without pre-processing or tokenizing identifiers.
The rationale is that such pre-processing should be part of the semantic representation.
For example, the NW string distance function aligns the characters of identifiers, and the FastText embeddings split identifiers into character n-grams, which may enable these techniques to reason about subtokens of an identifier.

\section{Results}
\label{sec:eval}

\subsection{RQ~1: Accuracy of Representing Semantic Relatedness}

\begin{figure*}
	\begin{subfigure}{.325\linewidth}
		\includegraphics[width=\linewidth]{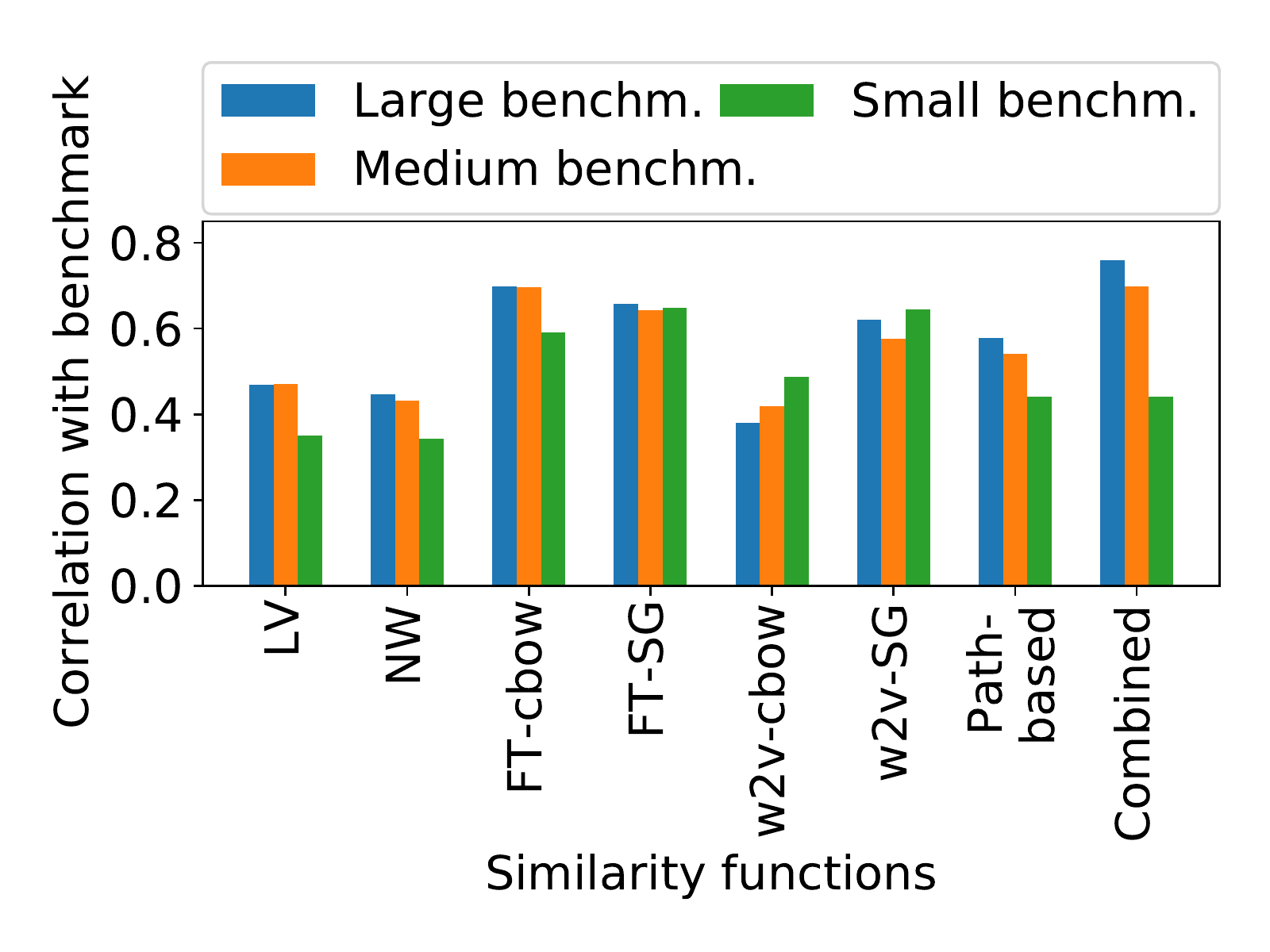}
		\caption{Relatedness.}
		\label{fig:correlations related}
	\end{subfigure}
	\hfill
	\begin{subfigure}{.325\linewidth}
		\includegraphics[width=\linewidth]{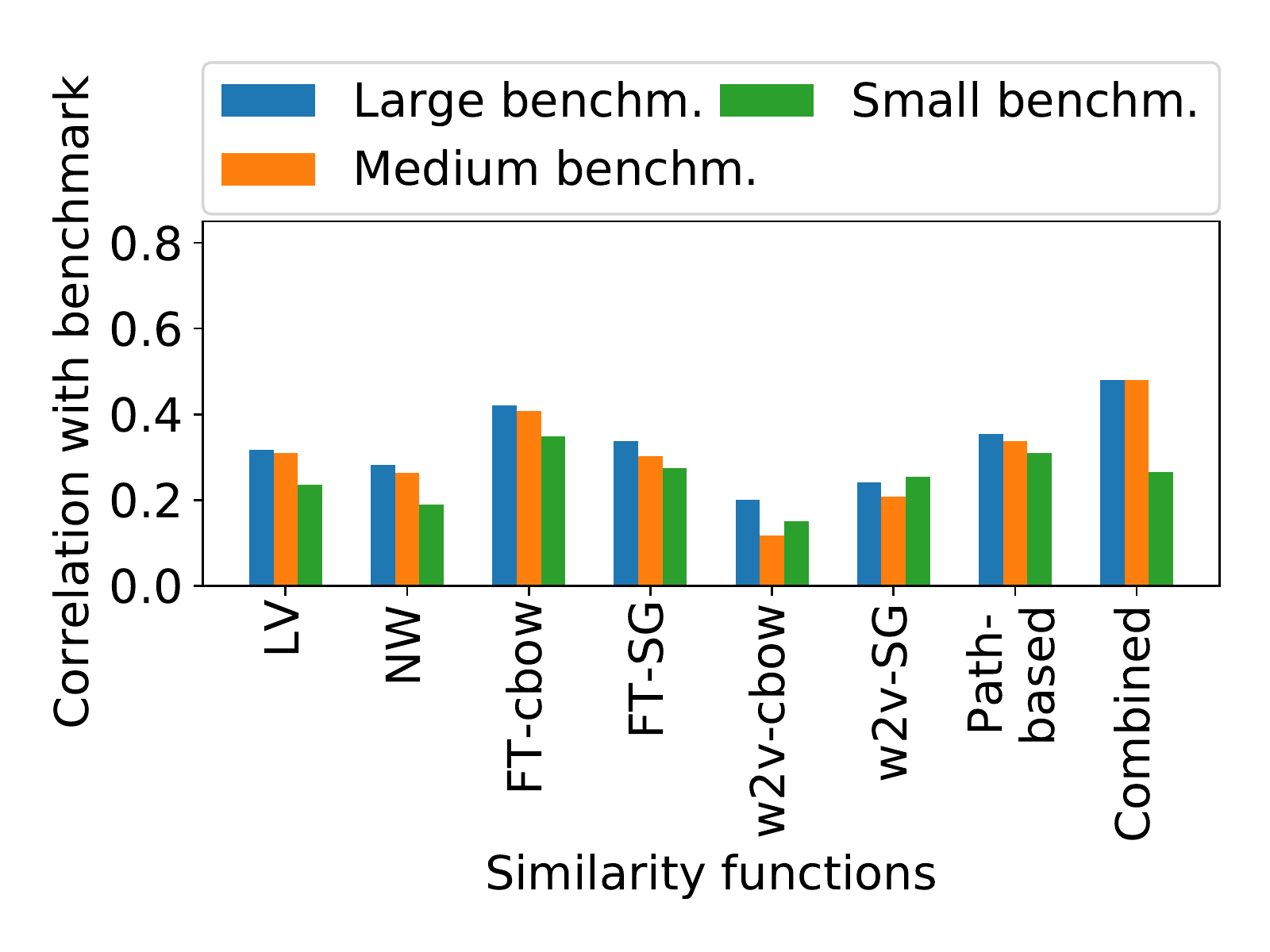}
		\caption{Similarity.}
		\label{fig:correlations similar}
	\end{subfigure}
	\hfill
	\begin{subfigure}{.325\linewidth}
		\includegraphics[width=\linewidth]{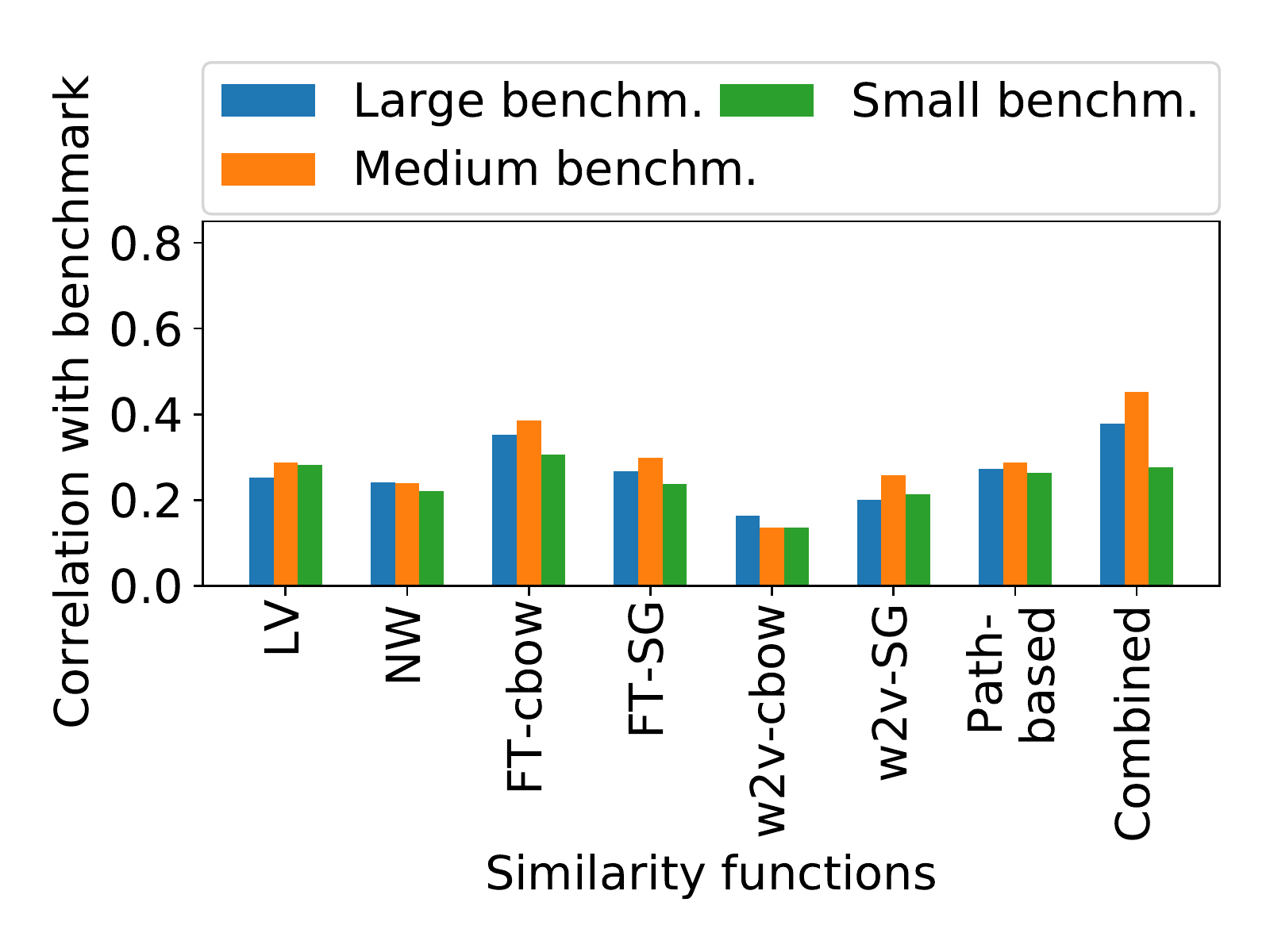}
		\caption{Contextual similarity.}
		\label{fig:correlations context}
	\end{subfigure}
	\caption{Correlations of embeddings and string distance functions with the small, medium, and large variants of the benchmark.}
	\label{fig:correlations}
\end{figure*}

The following addresses the question how accurately the studied techniques represent the relatedness of identifiers, i.e., the degree of association between the two identifiers.
Figure~\ref{fig:correlations related} shows the agreement of the evaluated semantic representations with the small, medium, and large variants of the relatedness benchmark in \name{}.
All techniques achieve relatively high levels of agreement, with correlations between 41\% and 74\%.
The neurally learned embeddings clearly outperform the string distance-based similarity functions (41-74\% vs.\ 46-49\%), showing that the effort of learning a semantic representation is worthwhile.
In particular, the learned embeddings match or even slightly exceed the inter-rater agreement, which is considered an upper bound of how strongly an embedding may correlate with a similarity-based benchmark~\cite{hill2015simlex}.
Comparing different embedding techniques with each other, we find that both FastText variants achieve higher scores than all other embeddings.
In contrast, despite using additional structural information of source code, path-based embeddings score only comparably to Word2vec.

A likely reason for the effectiveness of FastText is that it generalizes across lexically similar names by computing embeddings based on character n-grams of an identifier.
E.g., given the identifier \code{getIndex}, FastText computes its embedding based on embeddings for its various characters n-grams, such as \code{Index} and \code{Ind}, allowing the approach to generalize across lexically similar identifiers, such as \code{setIndex} or \code{ind}.

\subsection{RQ~2: Accuracy of Representing Semantic Similarity}

This research question is about the semantic similarity, i.e., the degree to which two identifiers have the same meaning.
Figure~\ref{fig:correlations similar} shows how much the studied semantic representations agree with the similarity benchmark in \name{}.
Overall, the figure shows a much lower agreement with the gold standard than for relatedness.
One explanation is that encoding semantic similarity is a harder task than encoding the less strict concept of relatedness.
Similar to relatedness, FT-cbow shows the strongest agreement, ranging between 35\% and 38\%.

The results of the contextual similarity task (Figure~\ref{fig:correlations context}) confirm the findings from the similarity task.
All studied techniques are less effective than for relatedness, and FT-cbow achieves the highest agreement with \name{}.

A perhaps surprising result is that string distance functions are roughly as effective as some of the learned embeddings and sometimes even outperform them.
The reason is that some semantically similar identifiers are also lexically similar, e.g., \code{len} and \code{length}.
One downside of string distance functions is that they miss synonymous identifiers, e.g., \code{count} and \code{total}.



\subsection{RQ~3: Strengths and Weaknesses of Existing Techniques}

To better understand the strengths and weaknesses of the studied semantic representations, we inspect various examples (Section~\ref{sec:examples}) and study interesting subsets of all identifier pairs in isolation (Section~\ref{sec:subsets}).

\subsubsection{Examples}
\label{sec:examples}

\begin{table*}[tb]
	\centering
	\setlength{\tabcolsep}{14pt}
	\caption[]{\label{tab:neighbor examples}Top-5 most similar identifiers by the FT-cbow and path-based models.}
	\begin{tabular}{@{}l|l|lllll@{}}
		\toprule
		Identifier & Embedding & \multicolumn{5}{c}{Nearest neighbors} \\
		\midrule
		
		substr & FT-cbow &substring&substrs&subst&substring1&substrCount\\
		& Path-based &substring&getInstanceProp&getPadding&getMinutes&floor\\
		\midrule		
		
		item& FT-cbow &itemNr&itemJ&itemL&itemI&itemAt\\
		& Path-based & entry&child&record&targ&nextElement\\
		\midrule
		
		count & FT-cbow &countTbl&countInt&countRTO&countsAsNum&countOne\\
		& Path-based &total&limit&minVal&exponent&rate\\
		\midrule
		
		rows & FT-cbow &rowOrRows&rowXs&rows\_l&rowsAr&rowIDs\\
		& Path-based &cols&cells&columns&tiles&items\\
		\midrule
		
		setInterval & FT-cbow &resetInterval&setTimeoutInterval&clearInterval&getInterval&retInterval\\
		& Path-based & clearInterval&assume&alert&nextTick&ReactTextComponent\\
		\midrule
		
		minText & FT-cbow &maxText&minLengthText&microsecText&maxLengthText&minuteText\\
		& Path-based & maxText&displayMsg&blankText&disableText&emptyText\\
		\midrule
		
		files & FT-cbow &filesObjs&filesGen&fileSets&extFiles&libFiles\\
		& Path-based & records&tasks&names&tiles&todos\\
		\midrule
		
		miny & FT-cbow &min\_y&minBy&minx&minPt&min\_z\\
		& Path-based & minx&ymin&dataMax&dataMin&ymax\\
		\bottomrule
	\end{tabular}
\end{table*}

To better understand why current embeddings sometimes fail to accurately represent similarities, Table~\ref{tab:neighbor examples} shows the most similar identifiers of selected identifiers according to the FT-cbow and path-based embeddings.
The examples illustrate two observations.
First, FastText, due to its use of n-grams~\cite{Bojanowski2017}, tends to cluster identifiers based on lexical commonalities.
While many lexically similar identifiers are also semantically similar, e.g., \code{substr} and \code{substring}, this approach misses other synonyms, e.g., \code{item} and \code{entry}.
Another downside is that lexical similarity may also establish wrong relationships.
For example, \code{substring} and \code{substrCount} represent different concepts, but FastText finds them to be highly similar.

Second, in contrast to FastText, path-based embeddings tend to cluster words based on the structural and syntactical contexts they occur in.
This approach helps the embeddings to identify synonyms despite their lexical differences, e.g., \code{count} and \code{total}, or \code{files} and \code{records}.
The downside is that it also clusters various related but not similar identifiers, e.g., \code{minText} and \code{maxText}, or \code{substr} and \code{getPadding}.
Some of these identifiers even have opposing meanings, e.g., \code{rows} and \code{cols}, which can mislead code analysis tools when reasoning about the semantics of code.

\subsubsection{Interesting Subsets of All Identifier Pairs}
\label{sec:subsets}

To better understand the strengths and weaknesses of semantic representations for specific kinds of identifiers, we analyze some interesting subsets of all identifier pairs in more detail.
We focus on four subsets:
\begin{itemize}
\item \emph{Abbreviations}. Pairs where at least one identifier is an abbreviation and where both identifiers refer to the same concept, e.g., \code{substr} and \code{substring}, or \code{cfg} and \code{conf}.
Since abbreviations are commonly used for concise source code, accurately reasoning about them is important.

\item \emph{Opposites}. Pairs where one identifier is the opposite of the other identifier, e.g., \code{xMin} and \code{xMax}, or \code{setInstanceProp} and \code{getInstanceProp}.
Since opposite identifiers often occur in similar contexts, they may be difficult to distinguish.

\item \emph{Synonyms}. Pairs that refer to the same concepts, e.g., \code{reset} and \code{clear}, or \code{emptyText} and \code{blankText}.
These identifiers often are lexically different but should be represented in a similar way.

\item \emph{Added subtoken}. Pairs where both identifiers are identical, except that one adds a subtoken to the other, e.g., \code{id} and \code{sessionid}, or \code{maxLine} and \code{maxLineLength}.

\item \emph{Tricky tokenization}. Pairs where at least one of the identifiers is composed of multiple subtokens but uses neither camel case nor snail case to combine subtokens, e.g., \code{touchmove} and \code{touchend}, or \code{newtext} and \code{content}.
This and the above subset are interesting because some semantic representations reason about subtokens of identifiers.
\end{itemize}
To extract pairs into these subsets, we inspect all 167 pairs from the small benchmark, which yields between 7 and 22 pairs per set.

\begin{figure}
\includegraphics[width=\linewidth]{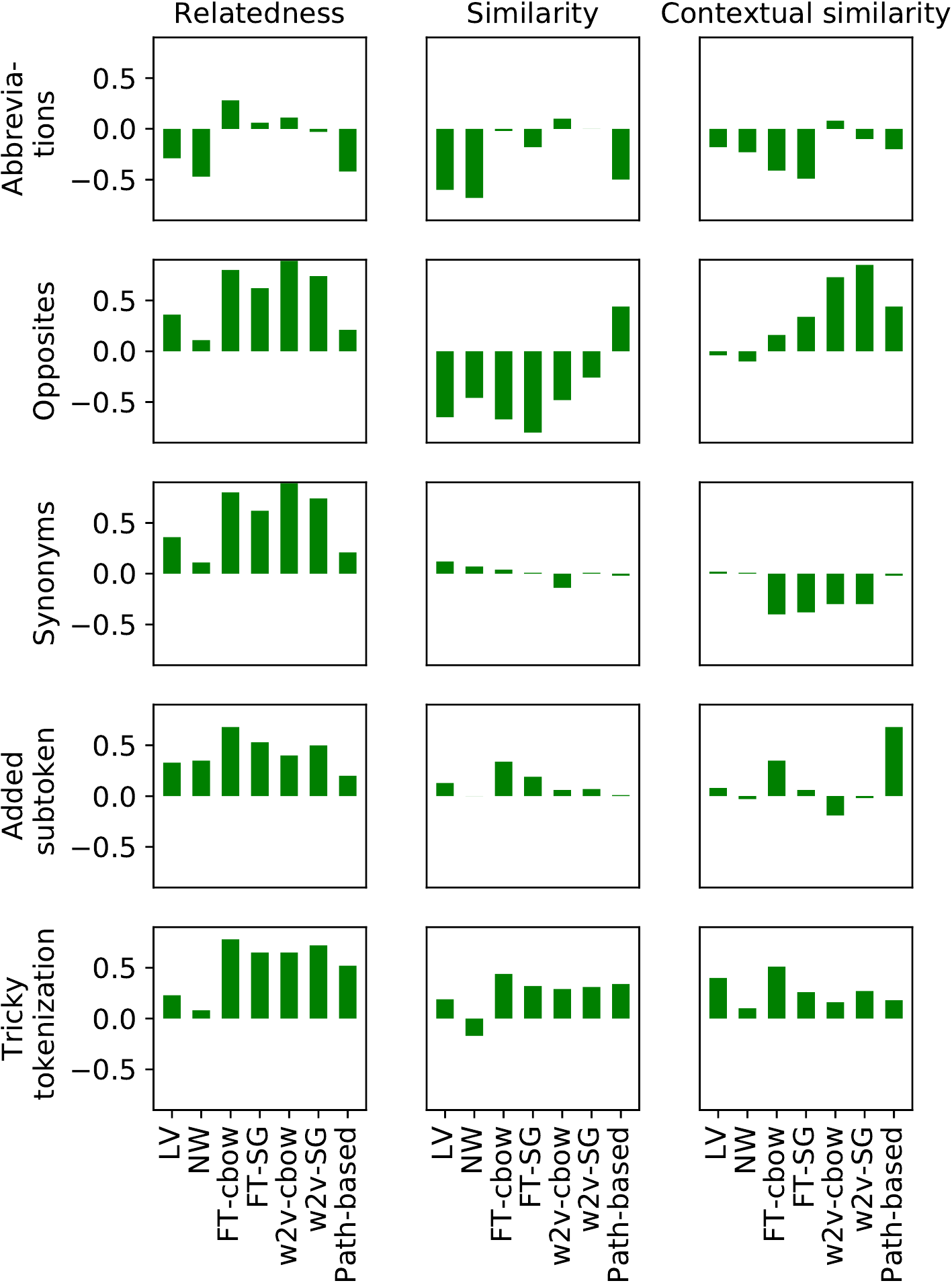}
\caption{Agreement and disagreement with the benchmark for different kinds of identifiers.}
\label{fig:subsets}
\end{figure}

Figure~\ref{fig:subsets} shows how much the different techniques agree or disagree with the benchmark for selected subsets.
As in Figure~\ref{fig:correlations}, each bar shows the Spearman rank correlation between the predicted similarities and the ground truth.
That is, higher values are better and negative values indicate a clear disagreement with the ground truth.

The results shows that all techniques are challenged by abbreviations, with more than half of the correlations being negative.
The poor performance for abbreviations can be attributed to the fact that fewer characters provide less information and that there may be many variants of the same name.
For opposites and synonyms, we find that most techniques, and in particular the learned embeddings, successfully represent the relatedness of these identifiers.
However, almost all techniques clearly fail to capture that opposite identifiers are not similar, as one cannot replace the other, and to capture that synonyms are similar.

For the subtoken-related subsets, we find that most techniques are challenged by pairs where one identifier adds a subtoken to the other, in particular, when reasoning about similarity.
One explanation is that identifiers with an added subtoken tend to be rather specialized, and hence, occur less frequent, which gives less training data to the learning-based techniques.
When being faced with identifiers that use non-obvious tokenization, most techniques, with the exception of Needleman-Wunsch, perform relatively well.
We attribute this result to the fact that techniques that reason about substrings of an identifier, such as FastText~\cite{Bojanowski2017}, do not rely on a specific tokenization approach, such as camel case or snail case, but instead consider character n-grams of the given identifier.

\subsection{RQ~4: Complementarity of Existing Techniques}
\label{sec:eval combination}

Our inspection of examples and of specific subsets of identifier pairs shows that different semantic representation techniques work well for different kinds of identifiers.
For example, some techniques work better for abbreviations than others.
Based on this observation, we hypothesize that the existing semantic representations complement each other.
If this hypothesis is correct, combining techniques in such a way that the most suitable set of techniques is used for a given pair of identifiers could represent similarities more accurately than any of the individual techniques.

To validate this hypothesis, we present an ensemble model that combines existing semantic representations.
The key idea is to train a model that predicts the similarity of two identifiers based on the similarity scores provided by the existing semantic representations.
To this end, the approach queries each of the seven techniques studied in this paper for a similarity score and provides these scores to the model.

To help the model decide what representations to favor for a given pair of identifiers, we also provide to the model a set of features that describe some properties of identifiers.
Given two identifiers, the features we consider are:
\begin{itemize}
\item The length of these identifiers.
\item The number of subtokens in each of the identifiers, based on snail case and camel case conventions.
\item The number of words among the subtokens that are not in an English dictionary. The rationale for this feature is to identify abbreviations, which usually are not dictionary words.
\end{itemize}

Given the seven similarity scores and the features, we train a model that takes the scores and features of a pair as an input, and then that predicts a similarity score for the pair.
We train the model in a supervised way, using the ground truth provided in \name{} as the labels for learning.
We use an off-the-shelf support vector machine model with the default hyperparameters provided by the underlying library\footnote{Class sklearn.svm.SVR from scikit-learn.}.

In practice, one would train the model with all pairs in our benchmark and then apply the trained model to new pairs.
To enable us to measure the effectiveness of the model, we here train it with all but one pair, and then apply the trained model to the left-out pair.
We repeat this step for each pair and use the score predicted by the model as the score of the combined technique.

Figure~\ref{fig:correlations} shows the results of the combined approach.
Combining different semantic representations clearly outperforms all existing techniques.
For example, for the large benchmark, the combined approach increases the relatedness, similarity, and contextual similarity of the best individual technique by 6\%, 19\%, and 5\%, respectively.
This result confirms our hypothesis that the existing techniques complement each other and shows the benefits of combining them.

\section{Discussion}


This section discusses some lessons learned from our study of semantic representations, along with ideas for addressing the current limitations in future work.

\paragraph{Neurally learned embeddings accurately represent the relatedness of identifiers}
Overall, all neural embeddings considered in our evaluation provide a high agreement with the ground truth provided by the relatedness scores in \name{}.
This result shows that embeddings are effective at assigning similar vector representations to identifiers that occur in the same application domain or that are associated in some other way.

\paragraph{No existing technique accurately represents the similarity of identifiers}
While the best available embeddings are highly effective at representing relatedness, none of the studied techniques reaches the same level of agreement for similarity.
In fact, even the best results in Figures~\ref{fig:correlations similar} and~\ref{fig:correlations context} (38\%) clearly stay beyond the inter-rater agreement of our benchmark (62\%), showing a huge potential for improvement.
For many applications of embeddings of identifiers, semantic similarity is crucial.
For example, techniques that suggest suitable variable or method names~\cite{Allamanis2015,alon2018general} aim for the name that is most similar, not only most related, to the concept represented by the variable or method.
Likewise, name-based analyses for finding programming errors~\cite{oopsla2018-DeepBugs} or variable misuses~\cite{Allamanis2017b} aim at identifying situations where the developer uses a wrong but perhaps related variable.
Improving the ability of semantic representations to accurately represent the similarity of identifiers will benefit these name-based analysis.

\paragraph{Neural embeddings generally outperform string distance functions}
Our results for both relatedness and similarity show that the best available neural embeddings outperform classical string distance functions.
For example, for the relatedness benchmark, the string distance functions achieve up to 49\% correlation, whereas embeddings achieve up to 74\% correlation.
For the similarity and contextual similarity benchmarks, the differences are smaller (32\% vs.\ 38\% for similarity, and 29\% vs.\ 36\% for contextual similarity), but still clearly visible.
These results suggest that name-based analyses are likely to benefit from using embeddings instead of string distance functions.

\paragraph{Opposite are challenging}
Inspecting examples of (in)accurately represented pairs of identifiers shows that identifiers that describe opposing concepts are particularly challenging for current semantic representations.
For example, both the FT-cbow and path-based embeddings assign similar vectors to \code{minText} and \code{maxText}, even though these identifiers are clearly not similar but only related.
Another example are the \code{setInterval} and \code{clearInterval} function names.
Table~\ref{tab:neighbor examples} shows these and other examples of this phenomenon.
Improving semantic representations to better distinguish identifiers with opposing meaning will benefit name-based analyses that, e.g., suggest method names~\cite{Allamanis2015} or refactorings of identifiers~\cite{Liu2019}.

\paragraph{Distinguishing singular and plural identifiers is particularly challenging}
Another challenge we observe while inspecting pairs of inaccurately represented pairs of identifiers is to distinguish identifiers of individual items from identifiers of collections of items.
For example, FT-cbow assigns very similar vectors to \code{substr} and \code{substrs} (Table~\ref{tab:neighbor examples}).
Such a conflation of singular and plural concepts may be misleading, e.g., for name-based analyses that predicts types~\cite{Xu2016,Hellendoorn2018,icse2019}.


\paragraph{Shared subword information may be misleading}
String distance functions and, to some extent, also subword-based embeddings, such as FastText, rely on the assumption that substrings shared by two identifiers increase the chance that the identifiers are semantically similar.
While a subword-based approach helps deal with the out-of-vocabulary problem~\cite{Babii2019}, it may also mislead the semantic representation.
For example, the FT-cbow embedding assigns similar vectors to \code{minText} and \code{minuteText}, as well as to \code{setInterval} and \code{clearInterval}, as these identifiers share subwords, even though the identifiers refer to clearly different concepts.

\paragraph{Expanding abbreviations may improve semantic representations}
The finding that practically all existing semantic representations have difficulties with abbreviations raises the question how to address this limitation.
One promising direction is to expand abbreviations into longer identifiers before querying for their relatedness or similarity to another identifier.
Several techniques for expanding identifiers have been proposed~\cite{DBLP:conf/icsm/CorazzaMM12,DBLP:conf/sigsoft/JiangLZ19,DBLP:conf/icsm/NewmanDAPKH19,DBLP:conf/icsm/LawrieB11,Lawrie2007}, which could possibly be used as a preprocessing step within semantic representations.

\paragraph{Different semantic representations complement each other}
The availability of different techniques for reasoning about the similarity of identifiers can be exploited by combining multiple such techniques.
Our ensemble model (Section~\ref{sec:eval combination}) shows the potential of combined approaches.

\section{Threats to Validity}

\subsection{Threats to Internal Validity}

Threats to internal validity are about factors that may influence our results.
The identifiers and the code examples associated with them may not be representative of other code.
To mitigate this threat, we gather data from a large and diverse code corpus, and we select identifiers that cover semantically similar and dissimilar pairs of identifiers (Section~\ref{sec:selection}).
The decision to perform our work with code written in a dynamically typed programming language, JavaScript, biases our results toward such languages.
The reason for focusing on a dynamically typed language is that such languages are the target of various name-based analyses~\cite{Xu2016,oopsla2018-DeepBugs,Hellendoorn2018,icse2019,Karampatsis2020,TypeWriter} and embedding techniques~\cite{alon2018general,alon2019code2vec}.

Some ratings gathered our surveys may be inaccurate, e.g., because participants may have misunderstood the instructions.
To mitigate this threat, we gather at least ten ratings per pair of identifiers and then carefully clean the ratings gathered by developers to remove noise and outliers (Section~\ref{sec:cleaning}).

\subsection{Threats to External Validity}

Threats to external validity are about factors that may influence the generalizability of our results.
One limitation is that \name{} focuses on individual identifiers only.
As a result, it is not clear to what extent our evaluation of semantic representations of identifiers allows for conclusions about representations at a larger granularity, e.g., of complex expressions, statements, or sequences of statements.
We focus on individual identifiers as they are the basic building blocks of code.
Recent work on improving name-based and learning-based bug detection~\cite{oopsla2018-DeepBugs} by aggregating identifiers in complex expressions suggests that improving embeddings for individual identifiers also benefits larger-scale code representations~\cite{Karampatsis2020}.

Another limitation is that other string distance functions or other embeddings may perform better or worse than those studied here.
We select semantic representations that have been used in past name-based analyses, as well as some recent embedding techniques that are state of the art in natural language processing (NLP).
By making \name{} publicly available, we enable others to evaluate future semantic representations.
As any benchmark, \name{} consists of a finite set of subjects, which may not be representative for all others.
The number of pairs of identifiers in the benchmark (Table~\ref{tab:benchmark size}) is in the same order of magnitude as that of word similarity benchmarks used in NLP~\cite{Wordsim-353,Schnabel2015,RG,miller1991contextual,hill2015simlex}.
Finally, we focus on JavaScript code, i.e., our findings may not generalize to identifiers in other languages.


Finally, different name-based analyses have different requirements on the semantic representations they build upon.
The tasks we present to survey participants may not represent all these requirements, and hence, a semantic representation may perform better or worse in a specific name-based analysis than \name{} suggests.

\section{Related Work}

\paragraph{Name-based Program Analysis}

Various analyses exploit the rich information provided by identifier names, e.g., to find bugs~\cite{issta2011,oopsla2017,oopsla2018-DeepBugs,DBLP:conf/sigsoft/KateOZEX18} and vulnerabilities~\cite{Harer2018a},
to mine specifications~\cite{Zhong2009a},
to infer types based on identifier names as implicit type hints~\cite{Xu2016,icse2019},
to predict the name of a method~\cite{Allamanis2015},
to complete partial code using a learned language model~\cite{DBLP:conf/acl/ChangPCR18},
to identify inappropriate names~\cite{Host2009},
to suggest more suitable names~\cite{Liu2019},
to resolve fully qualified type names of methods, variables, etc. in a given code snippet~\cite{Phan2018}, or to
map APIs between programming languages based on an embedding of code tokens~\cite{Nguyen2017}.
A systematic way of evaluating semantic representations of identifiers, as provided in this paper, helps in further exploiting the implicit knowledge encoded in identifiers, and hence will benefit name-based program analyses.

\paragraph{Embeddings of Identifiers}
Embeddings of identifiers are at the core of several code analysis tools.
A popular approach, e.g., for bug detection~\cite{oopsla2018-DeepBugs}, type prediction~\cite{icse2019}, or vulnerability detection~\cite{Harer2018a}, is applying Word2vec~\cite{word2veca,word2vecb} to token sequences, which corresponds to the Word2vec embedding evaluated in Section~\ref{sec:eval}.
\cite{White2016} train an RNN-based language model and extract its final hidden layer as an embedding of identifiers.
Chen et al.~\cite{chen2019literature} provide a more comprehensive survey of embeddings for source code.
Beyond learned embeddings, string distance functions are used in other name-based tools, e.g., for detecting bugs~\cite{issta2011,oopsla2017} or for inferring specifications~\cite{Zhong2009a}.
The quality of embeddings is crucial in these and other code analysis tools, and \name{} will help to improve the state of the art.

\paragraph{Embeddings of Programs}
Beyond embeddings of identifiers, there is work on embedding larger parts of a program.
One approach~\cite{Allamanis2015} uses a log-bilinear, neural language model~\cite{bengio2003neural} to predict the names of methods.
Other work embeds code based on graph neural networks~\cite{Allamanis2017b} or
sequence-based neural networks applied to paths through a graph representation of code~\cite{Ben-Nun2018,alon2019code2vec,Devlin2017,Henkel2018,DeFreez2018,Xu2017}.
Code2seq embeds code and then generates sequences of NL words~\cite{Alon2019a}.
For a broader overview and a detailed survey of learning-based software analysis, we refer the reader to \cite{NeuralSoftwareAnalysis} and \cite{Allamanis2018}, respectively.
To evaluate embeddings of programs, the COSET benchmark provides thousands of programs with semantic labels~\cite{Wang2019}.
Another study measures how effective pre-trained code2vec~\cite{alon2019code2vec} embeddings are for different downstream tasks~\cite{Kang2019}.
One conclusion from Kang et al.'s work~\cite{Kang2019} is that evaluating embeddings on a specific downstream task is insufficient, a problem we here address with a task-independent benchmark.
Both of the above~\cite{Wang2019,Kang2019} complement \name{} because the existing work is about entire programs, whereas \name{} is about identifiers.
Since identifiers are a basic building block of source code, a benchmark for improving embeddings of identifiers will eventually also benefit learning-based code analysis tools.

\paragraph{Benchmarks of Word Embeddings}
The NLP community has a long tradition of reasoning about the semantics of words.
In particular, that community has addressed the challenge of measuring how well a semantic representation of words matches actual relationships between words through a series of gold standards of words, focusing on either relatedness~\cite{Wordsim-353,MEN,Schnabel2015} or similarity~\cite{RG,miller1991contextual,hill2015simlex,gerz2016simverb} of words.
These gold standards define how similar two words are based on ratings by human judges, enabling an evaluation that measures how well an embedding reflects the human ratings.

Unfortunately, simply reusing these existing gold standards for identifiers in source code would be misleading.
One reason is that the vocabularies of natural languages and source code overlap only partially, because source code contains various terms and abbreviations not found in natural language texts.
Moreover, source code has a constantly growing vocabulary, as developers tend to quickly invent new identifiers, e.g., for newly emerging application domains~\cite{Babii2019}.
Finally, even words present in both natural languages and source code may differ in their meaning due to computer science-specific terms, e.g., ``float'' or ``string''.
This work is the first to address the need for a gold standard for identifiers in code.

\paragraph{Data Gathering}
Asking human raters how related or similar two words are was first proposed by ~\cite{RG} and then adopted by others~\cite{miller1991contextual,Wordsim-353,hill2015simlex,gerz2016simverb}.
Our direct survey also follows this methodology.
\cite{miller1991contextual} propose to gather judgments about contextual similarity by asking participants to choose a word to fill in a blank, an idea we adopt in our indirect survey.
To choose words and pairs of words, prior work relies on manual selection~\cite{RG}, pre-existing free association databases~\cite{hill2015simlex,gerz2016simverb}, e.g., USF~\cite{nelson2004university} or VerbNet~\cite{kipper2004extending,kipper2008large}, or cosine similarities according to pre-existing models~\cite{MEN}.
We follow the latter approach, as it minimizes human bias while covering a wide range of degrees of relatedness and similarity.

\paragraph{Inter-rater Agreement}
Validating and cleaning data gathered via crowd-sourcing based on the inter-rater agreement has been found effective in other crowd-sourced surveys~\cite{nowak2010reliable}.
Gold standards for natural language words reach an inter-rater agreement of 0.61~\cite{Wordsim-353} and 0.67~\cite{hill2015simlex}.
Our ``small'' dataset reaches similar levels of agreement, showing that the rates in \name{} represent a genuine human intuition.
As noted by \cite{hill2015simlex}, the inter-rater agreement also gives an upper bound of the expected correlation between the tested model and the gold standard.
Our results show that current models still leave plenty of room for improvement, especially w.r.t.\ similarity.

\section{Conclusion}

This paper presents the first benchmark for evaluating semantic representations of identifiers names, along with a study of current semantic representation techniques.
We compile thousands of ratings gathered from 500 developers into a benchmark that provides gold standard similarity scores representing the relatedness, similarity, and contextual similarity of identifiers.
Using \name{} to experimentally compare two string distance functions and five embedding techniques shows that these techniques differ significantly in their agreement with our gold standard.
The best available embeddings are effective at representing how related identifiers are.
However, all studied techniques show huge room for improvement in their ability to represent how similar identifiers are.
An in-depth study of different subsets of identifiers shows the specific strengths and weaknesses of current semantic representations, e.g., that most techniques are challenged by abbreviations, opposites, and the difference between singular and plural.
To exploit the complementarity of current techniques, we present an ensemble model that effectively combines them and clearly outperforms the best individual techniques.

Our work will help addressing the limitations of current semantic representations of identifiers.
Such progress will benefit downstream developer tools, in particular, name-based program analyses.
More broadly, improving semantic representations of identifiers will also contribute toward better learning-based program testing and analysis techniques.

\bibliographystyle{IEEEtran}
\bibliography{references}

\end{document}

%% file: lstlistings_javascript.tex
\lstdefinelanguage{JavaScript}{
  keywords={break, case, catch, continue, debugger, default, delete, do, else, finally, for, function, if, in, instanceof, new, return, switch, this, throw, try, typeof, var, void, while, with},
  morecomment=[l]{//},
  morecomment=[s]{/*}{*/},
  morestring=[b]',
  morestring=[b]",
  sensitive=true
}

\makeatletter
\newcommand\HL{%
   \gdef\lst@alloverstyle##1{%
     \fboxrule=0pt
     \fboxsep=0pt
     \colorbox{lightgray}{\strut##1}%
   }%
}

\newcommand\HLoff{%
   \xdef\lst@alloverstyle##1{##1}%
}

%% file: main.bbl
\begin{thebibliography}{10}
\providecommand{\url}[1]{#1}
\csname url@samestyle\endcsname
\providecommand{\newblock}{\relax}
\providecommand{\bibinfo}[2]{#2}
\providecommand{\BIBentrySTDinterwordspacing}{\spaceskip=0pt\relax}
\providecommand{\BIBentryALTinterwordstretchfactor}{4}
\providecommand{\BIBentryALTinterwordspacing}{\spaceskip=\fontdimen2\font plus
\BIBentryALTinterwordstretchfactor\fontdimen3\font minus
  \fontdimen4\font\relax}
\providecommand{\BIBforeignlanguage}[2]{{%
\expandafter\ifx\csname l@#1\endcsname\relax
\typeout{** WARNING: IEEEtran.bst: No hyphenation pattern has been}%
\typeout{** loaded for the language `#1'. Using the pattern for}%
\typeout{** the default language instead.}%
\else
\language=\csname l@#1\endcsname
\fi
#2}}
\providecommand{\BIBdecl}{\relax}
\BIBdecl

\bibitem{Butler2010}
S.~Butler, M.~Wermelinger, Y.~Yu, and H.~Sharp, ``Exploring the influence of
  identifier names on code quality: An empirical study,'' in \emph{European
  Conference on Software Maintenance and Reengineering (CSMR)}.\hskip 1em plus
  0.5em minus 0.4em\relax IEEE, 2010, pp. 156--165.

\bibitem{issta2011}
M.~Pradel and T.~R. Gross, ``Detecting anomalies in the order of equally-typed
  method arguments,'' in \emph{International Symposium on Software Testing and
  Analysis (ISSTA)}, 2011, pp. 232--242.

\bibitem{oopsla2017}
A.~Rice, E.~Aftandilian, C.~Jaspan, E.~Johnston, M.~Pradel, and
  Y.~Arroyo-Paredes, ``Detecting argument selection defects,'' in
  \emph{Conference on Object-Oriented Programming, Systems, Languages, and
  Applications (OOPSLA)}, 2017.

\bibitem{oopsla2018-DeepBugs}
\BIBentryALTinterwordspacing
M.~Pradel and K.~Sen, ``{DeepBugs}: A learning approach to name-based bug
  detection,'' \emph{{PACMPL}}, vol.~2, no. {OOPSLA}, pp. 147:1--147:25, 2018.
  [Online]. Available: \url{https://doi.org/10.1145/3276517}
\BIBentrySTDinterwordspacing

\bibitem{DBLP:conf/sigsoft/KateOZEX18}
\BIBentryALTinterwordspacing
S.~Kate, J.~Ore, X.~Zhang, S.~G. Elbaum, and Z.~Xu, ``Phys: probabilistic
  physical unit assignment and inconsistency detection,'' in \emph{Proceedings
  of the 2018 {ACM} Joint Meeting on European Software Engineering Conference
  and Symposium on the Foundations of Software Engineering, {ESEC/SIGSOFT}
  {FSE} 2018, Lake Buena Vista, FL, USA, November 04-09, 2018}, 2018, pp.
  563--573. [Online]. Available: \url{https://doi.org/10.1145/3236024.3236035}
\BIBentrySTDinterwordspacing

\bibitem{Zhong2009a}
H.~Zhong, L.~Zhang, T.~Xie, and H.~Mei, ``Inferring resource specifications
  from natural language {API} documentation,'' in \emph{International
  Conference on Automated Software Engineering (ASE)}, 2009, pp. 307--318.

\bibitem{Xu2016}
\BIBentryALTinterwordspacing
Z.~Xu, X.~Zhang, L.~Chen, K.~Pei, and B.~Xu, ``Python probabilistic type
  inference with natural language support,'' in \emph{Proceedings of the 24th
  {ACM} {SIGSOFT} International Symposium on Foundations of Software
  Engineering, {FSE} 2016, Seattle, WA, USA, November 13-18, 2016}, 2016, pp.
  607--618. [Online]. Available: \url{https://doi.org/10.1145/2950290.2950343}
\BIBentrySTDinterwordspacing

\bibitem{icse2019}
\BIBentryALTinterwordspacing
R.~S. Malik, J.~Patra, and M.~Pradel, ``{NL2Type}: {I}nferring {JavaScript}
  function types from natural language information,'' in \emph{Proceedings of
  the 41st International Conference on Software Engineering, {ICSE} 2019,
  Montreal, QC, Canada, May 25-31, 2019}, 2019, pp. 304--315. [Online].
  Available: \url{https://doi.org/10.1109/ICSE.2019.00045}
\BIBentrySTDinterwordspacing

\bibitem{Allamanis2015}
M.~Allamanis, E.~T. Barr, C.~Bird, and C.~A. Sutton, ``Suggesting accurate
  method and class names,'' in \emph{Proceedings of the 2015 10th Joint Meeting
  on Foundations of Software Engineering, {ESEC/FSE} 2015, Bergamo, Italy,
  August 30 - September 4, 2015}, 2015, pp. 38--49.

\bibitem{DBLP:conf/acl/ChangPCR18}
\BIBentryALTinterwordspacing
M.~R. Parvez, S.~Chakraborty, B.~Ray, and K.~Chang, ``Building language models
  for text with named entities,'' in \emph{Proceedings of the 56th Annual
  Meeting of the Association for Computational Linguistics, {ACL} 2018,
  Melbourne, Australia, July 15-20, 2018, Volume 1: Long Papers}, 2018, pp.
  2373--2383. [Online]. Available:
  \url{https://www.aclweb.org/anthology/P18-1221/}
\BIBentrySTDinterwordspacing

\bibitem{Host2009}
E.~W. H{\o}st and B.~M. {\O}stvold, ``Debugging method names,'' in
  \emph{European Conference on Object-Oriented Programming (ECOOP)}.\hskip 1em
  plus 0.5em minus 0.4em\relax Springer, 2009, pp. 294--317.

\bibitem{Liu2019}
\BIBentryALTinterwordspacing
K.~Liu, D.~Kim, T.~F. Bissyand{\'{e}}, T.~Kim, K.~Kim, A.~Koyuncu, S.~Kim, and
  Y.~L. Traon, ``Learning to spot and refactor inconsistent method names,'' in
  \emph{Proceedings of the 41st International Conference on Software
  Engineering, {ICSE} 2019, Montreal, QC, Canada, May 25-31, 2019}, 2019, pp.
  1--12. [Online]. Available: \url{https://dl.acm.org/citation.cfm?id=3339507}
\BIBentrySTDinterwordspacing

\bibitem{Butler2011}
S.~Butler, M.~Wermelinger, Y.~Yu, and H.~Sharp, ``Improving the tokenisation of
  identifier names,'' in \emph{European Conference on Object-Oriented
  Programming (ECOOP)}.\hskip 1em plus 0.5em minus 0.4em\relax Springer, 2011,
  pp. 130--154.

\bibitem{Jiang2018}
Y.~Jiang, H.~Liu, J.~Q. Zhu, and L.~Zhang, ``Automatic and accurate expansion
  of abbreviations in parameters,'' \emph{IEEE Transactions on Software
  Engineering}, 2018.

\bibitem{icse2016-names}
H.~Liu, Q.~Liu, C.-A. Staicu, M.~Pradel, and Y.~Luo, ``Nomen est omen:
  Exploring and exploiting similarities between argument and parameter names,''
  in \emph{International Conference on Software Engineering (ICSE)}, 2016, pp.
  1063--1073.

\bibitem{NeuralSoftwareAnalysis}
\BIBentryALTinterwordspacing
M.~Pradel and S.~Chandra, ``Neural software analysis,'' \emph{CoRR}, vol.
  abs/2011.07986, 2020. [Online]. Available:
  \url{https://arxiv.org/abs/2011.07986}
\BIBentrySTDinterwordspacing

\bibitem{alon2018general}
U.~Alon, M.~Zilberstein, O.~Levy, and E.~Yahav, ``A general path-based
  representation for predicting program properties,'' in \emph{ACM SIGPLAN
  Notices}, vol.~53, no.~4.\hskip 1em plus 0.5em minus 0.4em\relax ACM, 2018,
  pp. 404--419.

\bibitem{Nguyen2017}
T.~D. Nguyen, A.~T. Nguyen, H.~D. Phan, and T.~N. Nguyen, ``Exploring {API}
  embedding for {API} usages and applications,'' in \emph{Proceedings of the
  39th International Conference on Software Engineering, {ICSE} 2017, Buenos
  Aires, Argentina, May 20-28, 2017}, 2017, pp. 438--449.

\bibitem{Hellendoorn2018}
V.~Hellendoorn, C.~Bird, E.~T. Barr, and M.~Allamanis, ``Deep learning type
  inference,'' in \emph{FSE}, 2018.

\bibitem{miller1991contextual}
G.~A. Miller and W.~G. Charles, ``Contextual correlates of semantic
  similarity,'' \emph{Language and cognitive processes}, vol.~6, no.~1, pp.
  1--28, 1991.

\bibitem{Allamanis2017b}
\BIBentryALTinterwordspacing
M.~Allamanis, M.~Brockschmidt, and M.~Khademi, ``Learning to represent programs
  with graphs,'' \emph{CoRR}, vol. abs/1711.00740, 2017. [Online]. Available:
  \url{http://arxiv.org/abs/1711.00740}
\BIBentrySTDinterwordspacing

\bibitem{Bojanowski2017}
\BIBentryALTinterwordspacing
P.~Bojanowski, E.~Grave, A.~Joulin, and T.~Mikolov, ``Enriching word vectors
  with subword information,'' \emph{{TACL}}, vol.~5, pp. 135--146, 2017.
  [Online]. Available:
  \url{https://transacl.org/ojs/index.php/tacl/article/view/999}
\BIBentrySTDinterwordspacing

\bibitem{alon2019code2vec}
U.~Alon, M.~Zilberstein, O.~Levy, and E.~Yahav, ``code2vec: Learning
  distributed representations of code,'' \emph{Proceedings of the ACM on
  Programming Languages}, vol.~3, no. POPL, p.~40, 2019.

\bibitem{raychev2016learning_150KJavaScript}
V.~Raychev, P.~Bielik, M.~Vechev, and A.~Krause, ``Learning programs from noisy
  data,'' in \emph{ACM SIGPLAN Notices}, vol.~51, no.~1.\hskip 1em plus 0.5em
  minus 0.4em\relax ACM, 2016, pp. 761--774.

\bibitem{MEN}
E.~Bruni, N.-K. Tran, and M.~Baroni, ``Multimodal distributional semantics,''
  \emph{Journal of Artificial Intelligence Research}, vol.~49, pp. 1--47, 2014.

\bibitem{word2veca}
T.~Mikolov, K.~Chen, G.~Corrado, and J.~Dean, ``Efficient estimation of word
  representations in vector space,'' \emph{arXiv preprint arXiv:1301.3781},
  2013.

\bibitem{Raychev2016a}
V.~Raychev, P.~Bielik, and M.~Vechev, ``Probabilistic model for code with
  decision trees,'' in \emph{OOPSLA}, 2016.

\bibitem{Allamanis2013}
M.~Allamanis and C.~A. Sutton, ``Mining source code repositories at massive
  scale using language modeling,'' in \emph{Proceedings of the 10th Working
  Conference on Mining Software Repositories, {MSR} '13, San Francisco, CA,
  USA, May 18-19, 2013}, 2013, pp. 207--216.

\bibitem{kittur2008crowdsourcing}
A.~Kittur, E.~H. Chi, and B.~Suh, ``Crowdsourcing user studies with mechanical
  turk,'' in \emph{Proceedings of the SIGCHI conference on human factors in
  computing systems}, 2008, pp. 453--456.

\bibitem{nowak2010reliable}
S.~Nowak and S.~R{\"u}ger, ``How reliable are annotations via crowdsourcing: a
  study about inter-annotator agreement for multi-label image annotation,'' in
  \emph{Proceedings of the international conference on Multimedia information
  retrieval}, 2010, pp. 557--566.

\bibitem{hill2015simlex}
F.~Hill, R.~Reichart, and A.~Korhonen, ``Simlex-999: Evaluating semantic models
  with (genuine) similarity estimation,'' \emph{Computational Linguistics},
  vol.~41, no.~4, pp. 665--695, 2015.

\bibitem{zhelezniak2019correlation}
V.~Zhelezniak, A.~Savkov, A.~Shen, and N.~Y. Hammerla, ``Correlation
  coefficients and semantic textual similarity,'' \emph{arXiv preprint
  arXiv:1905.07790}, 2019.

\bibitem{needleman1970general}
S.~B. Needleman and C.~D. Wunsch, ``A general method applicable to the search
  for similarities in the amino acid sequence of two proteins,'' \emph{Journal
  of molecular biology}, vol.~48, no.~3, pp. 443--453, 1970.

\bibitem{word2vecb}
T.~Mikolov, I.~Sutskever, K.~Chen, G.~S. Corrado, and J.~Dean, ``Distributed
  representations of words and phrases and their compositionality,'' in
  \emph{Advances in neural information processing systems}, 2013, pp.
  3111--3119.

\bibitem{Babii2019}
\BIBentryALTinterwordspacing
H.~Babii, A.~Janes, and R.~Robbes, ``Modeling vocabulary for big code machine
  learning,'' \emph{CoRR}, 2019. [Online]. Available:
  \url{https://arxiv.org/abs/1904.01873}
\BIBentrySTDinterwordspacing

\bibitem{DBLP:conf/icsm/CorazzaMM12}
\BIBentryALTinterwordspacing
A.~Corazza, S.~D. Martino, and V.~Maggio, ``{LINSEN:} an efficient approach to
  split identifiers and expand abbreviations,'' in \emph{28th {IEEE}
  International Conference on Software Maintenance, {ICSM} 2012, Trento, Italy,
  September 23-28, 2012}.\hskip 1em plus 0.5em minus 0.4em\relax {IEEE}
  Computer Society, 2012, pp. 233--242. [Online]. Available:
  \url{https://doi.org/10.1109/ICSM.2012.6405277}
\BIBentrySTDinterwordspacing

\bibitem{DBLP:conf/sigsoft/JiangLZ19}
\BIBentryALTinterwordspacing
Y.~Jiang, H.~Liu, and L.~Zhang, ``Semantic relation based expansion of
  abbreviations,'' in \emph{Proceedings of the {ACM} Joint Meeting on European
  Software Engineering Conference and Symposium on the Foundations of Software
  Engineering, {ESEC/SIGSOFT} {FSE} 2019, Tallinn, Estonia, August 26-30,
  2019}, M.~Dumas, D.~Pfahl, S.~Apel, and A.~Russo, Eds.\hskip 1em plus 0.5em
  minus 0.4em\relax {ACM}, 2019, pp. 131--141. [Online]. Available:
  \url{https://doi.org/10.1145/3338906.3338929}
\BIBentrySTDinterwordspacing

\bibitem{DBLP:conf/icsm/NewmanDAPKH19}
\BIBentryALTinterwordspacing
C.~D. Newman, M.~J. Decker, R.~S. Alsuhaibani, A.~Peruma, D.~Kaushik, and
  E.~Hill, ``An empirical study of abbreviations and expansions in software
  artifacts,'' in \emph{2019 {IEEE} International Conference on Software
  Maintenance and Evolution, {ICSME} 2019, Cleveland, OH, USA, September 29 -
  October 4, 2019}.\hskip 1em plus 0.5em minus 0.4em\relax {IEEE}, 2019, pp.
  269--279. [Online]. Available: \url{https://doi.org/10.1109/ICSME.2019.00040}
\BIBentrySTDinterwordspacing

\bibitem{DBLP:conf/icsm/LawrieB11}
\BIBentryALTinterwordspacing
D.~J. Lawrie and D.~W. Binkley, ``Expanding identifiers to normalize source
  code vocabulary,'' in \emph{{IEEE} 27th International Conference on Software
  Maintenance, {ICSM} 2011, Williamsburg, VA, USA, September 25-30,
  2011}.\hskip 1em plus 0.5em minus 0.4em\relax {IEEE} Computer Society, 2011,
  pp. 113--122. [Online]. Available:
  \url{https://doi.org/10.1109/ICSM.2011.6080778}
\BIBentrySTDinterwordspacing

\bibitem{Lawrie2007}
D.~Lawrie, H.~Feild, and D.~Binkley, ``Extracting meaning from abbreviated
  identifiers,'' in \emph{Working Conference on Source Code Analysis and
  Manipulation (SCAM)}.\hskip 1em plus 0.5em minus 0.4em\relax IEEE, 2007, pp.
  213--222.

\bibitem{Karampatsis2020}
\BIBentryALTinterwordspacing
R.-M. Karampatsis and C.~Sutton, ``Scelmo: Source code embeddings from language
  models,'' 2020. [Online]. Available:
  \url{https://openreview.net/pdf?id=ryxnJlSKvr}
\BIBentrySTDinterwordspacing

\bibitem{TypeWriter}
\BIBentryALTinterwordspacing
M.~Pradel, G.~Gousios, J.~Liu, and S.~Chandra, ``Typewriter: Neural type
  prediction with search-based validation,'' in \emph{{ESEC/FSE} '20: 28th
  {ACM} Joint European Software Engineering Conference and Symposium on the
  Foundations of Software Engineering, Virtual Event, USA, November 8-13,
  2020}, 2020, pp. 209--220. [Online]. Available:
  \url{https://doi.org/10.1145/3368089.3409715}
\BIBentrySTDinterwordspacing

\bibitem{Wordsim-353}
L.~Finkelstein, E.~Gabrilovich, Y.~Matias, E.~Rivlin, Z.~Solan, G.~Wolfman, and
  E.~Ruppin, ``Placing search in context: The concept revisited,'' \emph{ACM
  Transactions on information systems}, vol.~20, no.~1, pp. 116--131, 2002.

\bibitem{Schnabel2015}
\BIBentryALTinterwordspacing
T.~Schnabel, I.~Labutov, D.~M. Mimno, and T.~Joachims, ``Evaluation methods for
  unsupervised word embeddings,'' in \emph{Proceedings of the 2015 Conference
  on Empirical Methods in Natural Language Processing, {EMNLP} 2015, Lisbon,
  Portugal, September 17-21, 2015}, 2015, pp. 298--307. [Online]. Available:
  \url{http://aclweb.org/anthology/D/D15/D15-1036.pdf}
\BIBentrySTDinterwordspacing

\bibitem{RG}
H.~Rubenstein and J.~B. Goodenough, ``Contextual correlates of synonymy,''
  \emph{Communications of the ACM}, vol.~8, no.~10, pp. 627--633, 1965.

\bibitem{Kang2019}
H.~J. Kang, T.~F. Bissyand\'e, and D.~Lo, ``Assessing the generalizability of
  code2vec token embeddings,'' in \emph{ASE}, 2019.

\bibitem{Harer2018a}
\BIBentryALTinterwordspacing
J.~A. Harer, L.~Y. Kim, R.~L. Russell, O.~Ozdemir, L.~R. Kosta, A.~Rangamani,
  L.~H. Hamilton, G.~I. Centeno, J.~R. Key, P.~M. Ellingwood, M.~W. McConley,
  J.~M. Opper, S.~P. Chin, and T.~Lazovich, ``Automated software vulnerability
  detection with machine learning,'' \emph{CoRR}, vol. abs/1803.04497, 2018.
  [Online]. Available: \url{http://arxiv.org/abs/1803.04497}
\BIBentrySTDinterwordspacing

\bibitem{Phan2018}
\BIBentryALTinterwordspacing
H.~Phan, H.~A. Nguyen, N.~M. Tran, L.~H. Truong, A.~T. Nguyen, and T.~N.
  Nguyen, ``Statistical learning of {API} fully qualified names in code
  snippets of online forums,'' in \emph{Proceedings of the 40th International
  Conference on Software Engineering, {ICSE} 2018, Gothenburg, Sweden, May 27 -
  June 03, 2018}, 2018, pp. 632--642. [Online]. Available:
  \url{http://doi.acm.org/10.1145/3180155.3180230}
\BIBentrySTDinterwordspacing

\bibitem{White2016}
M.~White, M.~Tufano, C.~Vendome, and D.~Poshyvanyk, ``Deep learning code
  fragments for code clone detection,'' in \emph{ASE}, 2016, pp. 87--98.

\bibitem{chen2019literature}
Z.~Chen and M.~Monperrus, ``A literature study of embeddings on source code,''
  \emph{arXiv preprint arXiv:1904.03061}, 2019.

\bibitem{bengio2003neural}
Y.~Bengio, R.~Ducharme, P.~Vincent, and C.~Jauvin, ``A neural probabilistic
  language model,'' \emph{Journal of machine learning research}, vol.~3, no.
  Feb, pp. 1137--1155, 2003.

\bibitem{Ben-Nun2018}
\BIBentryALTinterwordspacing
T.~Ben{-}Nun, A.~S. Jakobovits, and T.~Hoefler, ``Neural code comprehension:
  {A} learnable representation of code semantics,'' \emph{CoRR}, vol.
  abs/1806.07336, 2018. [Online]. Available:
  \url{http://arxiv.org/abs/1806.07336}
\BIBentrySTDinterwordspacing

\bibitem{Devlin2017}
\BIBentryALTinterwordspacing
J.~Devlin, J.~Uesato, R.~Singh, and P.~Kohli, ``Semantic code repair using
  neuro-symbolic transformation networks,'' \emph{CoRR}, vol. abs/1710.11054,
  2017. [Online]. Available: \url{http://arxiv.org/abs/1710.11054}
\BIBentrySTDinterwordspacing

\bibitem{Henkel2018}
J.~Henkel, S.~K. Lahiri, B.~Liblit, and T.~W. Reps, ``Code vectors:
  understanding programs through embedded abstracted symbolic traces,'' in
  \emph{Proceedings of the 2018 {ACM} Joint Meeting on European Software
  Engineering Conference and Symposium on the Foundations of Software
  Engineering, {ESEC/SIGSOFT} {FSE} 2018, Lake Buena Vista, FL, USA, November
  04-09, 2018}, 2018, pp. 163--174.

\bibitem{DeFreez2018}
D.~DeFreez, A.~V. Thakur, and C.~Rubio{-}Gonz{\'{a}}lez, ``Path-based function
  embedding and its application to specification mining,'' \emph{CoRR}, vol.
  abs/1802.07779, 2018.

\bibitem{Xu2017}
X.~Xu, C.~Liu, Q.~Feng, H.~Yin, L.~Song, and D.~Song, ``Neural network-based
  graph embedding for cross-platform binary code similarity detection,'' in
  \emph{CCS}, 2017, pp. 363--376.

\bibitem{Alon2019a}
\BIBentryALTinterwordspacing
U.~Alon, S.~Brody, O.~Levy, and E.~Yahav, ``code2seq: Generating sequences from
  structured representations of code,'' in \emph{7th International Conference
  on Learning Representations, {ICLR} 2019, New Orleans, LA, USA, May 6-9,
  2019}, 2019. [Online]. Available:
  \url{https://openreview.net/forum?id=H1gKYo09tX}
\BIBentrySTDinterwordspacing

\bibitem{Allamanis2018}
M.~Allamanis, E.~T. Barr, P.~Devanbu, and C.~Sutton, ``A survey of machine
  learning for big code and naturalness,'' \emph{ACM Computing Surveys (CSUR)},
  vol.~51, no.~4, p.~81, 2018.

\bibitem{Wang2019}
\BIBentryALTinterwordspacing
K.~Wang and M.~Christodorescu, ``Coset: A benchmark for evaluating neural
  program embeddings,'' \emph{CoRR}, 2019. [Online]. Available:
  \url{https://arxiv.org/abs/1905.11445}
\BIBentrySTDinterwordspacing

\bibitem{gerz2016simverb}
D.~Gerz, I.~Vuli{\'c}, F.~Hill, R.~Reichart, and A.~Korhonen, ``Simverb-3500: A
  large-scale evaluation set of verb similarity,'' \emph{arXiv preprint
  arXiv:1608.00869}, 2016.

\bibitem{nelson2004university}
D.~L. Nelson, C.~L. McEvoy, and T.~A. Schreiber, ``The university of south
  florida free association, rhyme, and word fragment norms,'' \emph{Behavior
  Research Methods, Instruments, \& Computers}, vol.~36, no.~3, pp. 402--407,
  2004.

\bibitem{kipper2004extending}
K.~Kipper, B.~Snyder, and M.~Palmer, ``Extending a verb-lexicon using a
  semantically annotated corpus.'' in \emph{LREC}, 2004.

\bibitem{kipper2008large}
K.~Kipper, A.~Korhonen, N.~Ryant, and M.~Palmer, ``A large-scale classification
  of english verbs,'' \emph{Language Resources and Evaluation}, vol.~42, no.~1,
  pp. 21--40, 2008.

\end{thebibliography}
